# Evolutionary Co-Design of Morphology and Control of Soft Tensegrity Modular Robots with Programmable Stiffness

Davide Zappetti*
Laboratory of Intelligent Systems,
Ecole polytechnique fédérale de Lausanne (EPFL)
1015, Lausanne, Switzerland
davide.zappetti@epfl.ch

Jean Marc Bejjani*
Laboratory of Intelligent Systems,
Ecole polytechnique fédérale de Lausanne (EPFL)
1015, Lausanne, Switzerland
jean.bejjani@epfl.ch

Dario Floreano
Laboratory of Intelligent Systems,
Ecole polytechnique fédérale de Lausanne (EPFL)
1015, Lausanne, Switzerland
dario.floreano@epfl.ch

*D.Z. and J.M.B. contributed equally to this work

*Abstract*— Tensegrity structures are lightweight, can undergo large deformations, and have outstanding robustness capabilities. These unique properties inspired roboticists to investigate their use. However, the morphological design, control, assembly, and actuation of tensegrity robots are still difficult tasks. Moreover, the stiffness of tensegrity robots is still an underestimated design parameter. In this article, we propose to use easy to assemble, actuated tensegrity modules and body-brain co-evolution to design soft tensegrity modular robots. Moreover, we prove the importance of tensegrity robots stiffness showing how the evolution suggests a different morphology, control and locomotion strategy according to the modules stiffness.

## I. Introduction

Tensegrity structures are lightweight, can undergo large deformations, and have outstanding robustness capabilities [1]. These unique properties have inspired roboticists [4] to investigate their use. Tensegrity robots were initially suggested and investigated by Haller et al. [5], while Rieffel et al. suggested the use of tensegrities to achieve "mechanical intelligence" in artificial agents [6]. More recently the use of tensegrity has been suggested and investigated in the field of soft robotics [7]. Indeed, when tensegrity structures include elastic cables, they show high deformability, displaying a behavior akin to that of soft materials.

The morphological design, control, assembly, and actuation of tensegrity robots are still difficult tasks. Indeed, most tensegrity robots so far have simple polyhedral tensegrity bodies (i.e. tensegrity structure resembling polyhedral convex solids) [8-10] or fixed mechanical properties (e.g. stiffness of the cables) [11-15]. These morphology designs could be suboptimal according to the task or the goal of the robot. Therefore, in our recent work [17], we proposed a modular approach, using simple and easy to assemble tensegrity structures with programmable stiffness and actuation as building blocks for constructing larger and more complex robotic morphologies with programmed stiffness. However, how to properly connect, programme stiffness and control each module, according to a specific task, is still an open challenge.

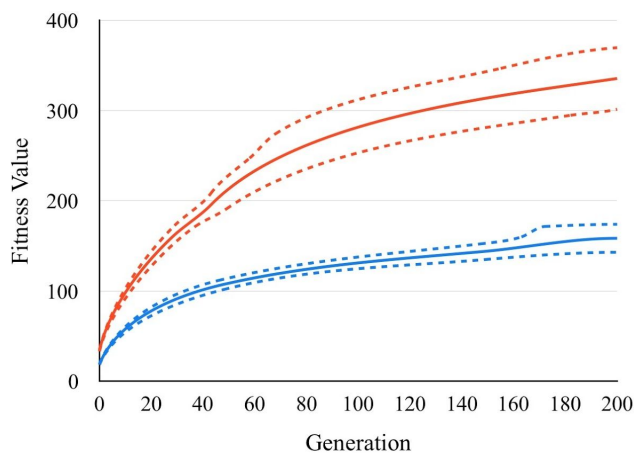

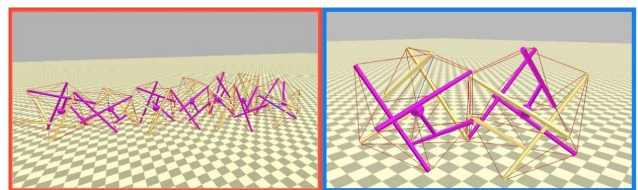

Example of High Stiffness evolved Robot | Example of Low Stiffness evolved Robot

Fig. 1. Body-brain co-evolution fitness versus generations graph for two set of experiments. In blue the set with only low stiffness modules allowed and in orange with only high stiffness modules.

Evolutionary algorithms have been shown to be a promising approach to design complex soft robot morphologies, their stiffness and control [18-19]. So far evolutionary algorithms have been successfully applied to evolve tensegrity morphologies without control [20] or control strategies of fixed tensegrity morphology [21-24].

In this work we apply evolutionary algorithms to perform successful co-evolution of body and brain (e.g. morphology and control) of tensegrity modular robots, and we show the importance of determining the appropriate stiffness in tensegrity soft robots given a specific task. Indeed, we show how changing the stiffness of the modules -remaining in a soft domain- affects not only the body shape and control of the robot solution, but also evolves different locomotion strategies.

## II. Design and Manufacturing of the modules

In this section we describe the design of the tensegrity modules with programmable stiffness and actuation that were simulated to perform body-brain co-evolution. Moreover, we describe their manufacturing strategy and easy assembly process.

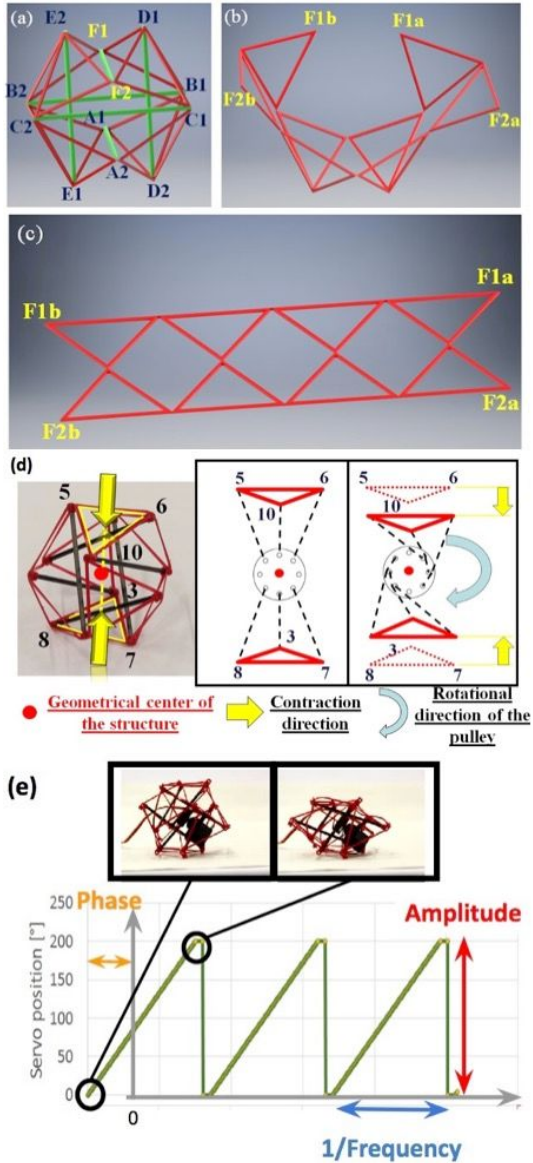

Fig. 2. (a) 3D CAD model of a tensegrity icosahedron. The two vertices disconnected to unfold the cable's network are marked in yellow. (b-c) The unfolding sequence obtained rotating the triangles around the joints in the vertices, in yellow are marked the 4 nodes that will be overlapped to generate vertices F1 and F2. (d) Actuation strategy. (e) Sawtooth control signal of an actuated module. Phase, frequency and amplitude are the design parameters.

### A. Design

The well known icosahedron tensegrity structure [1] has been selected as the main module of our modular tensegrity robot [17]. Three main criteria led to the choice [17]. Firstly, the ability to deform along a direction orthogonal to any of the 8 triangles that it displays on its external surface, allowing deformations in three-dimensional space. Moreover, the same 8 faces can be used to connect the modules to each other as shown in our previous work [17]. Secondly, the icosahedron tensegrity is one of the symmetric structures composed of the lowest amount of struts and cables, which eases the manufacturing and assembly. Finally, it has an inner cubic volume that is not crossed by any cable or strut to allow placement and protection of a useful payload [17] (i.e. an actuator)(fig 2a).

The orientation of connected modules, and which of the 8 faces these are connected to are the evolved morphological parameters considered in the body-brain evolution design described in section III.

### B. Actuation

An actuation mechanism was added to the main module in order to control its contraction along one of the 8 orthogonal directions to the connective faces. In [17] we proposed to use a tendon-driven contractive system operated by a servo-motor that is strategically placed to avoid rigid connection between struts, thus preserving the tensegrity nature of the structure and its deformability [1] (fig 2d). The direction of the actuation for each module (e.g one of the 8 orthogonal directions) is another evolved morphological parameter in the body-brain evolution design described in section III.

### C. Control

The servo-motor signal in each module can be individually tuned. The signal given is a sawtooth that can be modulated in terms of amplitude, frequency and offset. These three control parameters are evolved for each module in the body-brain evolution design described in section III (fig. 2e).

### D. Manufacturing

In our last work [17] we also proposed a new manufacturing strategy based on 3D printing. Instead of manufacturing every single component separately, we proposed to manufacture all cables as a single flat network that can be folded into a three-dimensional structure and filled with struts (fig. 2a-c) Moreover, the 3D CAD for the printing can be easily modified to change cables parameters and program the module stiffness. Such approach represents an important feature in terms of scalability for a modular robot design. Once assembled, the modules can be easily connected by mechanical latching at the vertices of the connected faces [17].

## III. Co-evolution of body and brain of the tensegrity modular robots with programmed stiffness

In this section we describe how we perform simulations and body-brain evolution of modular tensegrity robots with programmed stiffness based on our design.

For co-evolving body-brain we developed a Modular Tensegrity evolution platform named: "TensSoft", which is built with additional libraries on top of the NASA Tensegrity Robotics Toolkit (NTRT) [25]. The NTRT allows to simulate modular tensegrity robots based on our current hardware with a good reality matching [21].

"TensSoft" implements an easy to use input structure definition for modular tensegrity robots based on our current modular design. The input allows the user to define any robot with a tree-shaped structure.

We also implemented an evolution algorithm layer called "TensSoft Evolution" that can evaluate and evolve the body and brain of modular robots in order to optimize a certain behavior with a fitness function. For that layer, we used the "GAlib 2.4" library for genetic algorithms developed at MIT [26].

The evolution layer currently uses a direct encoding where each robot is defined by two types of genes; morphological genes that define the morphology (e.g the number of modules) and control genes that define the open loop input signals for each independent module of the structure. For this experiment, it defines for each module the frequency, amplitude, phase and the orientation of it's actuation.

We use a binary genome from the "GAlib" for the encoding and the built-in Steady State Genetic Algorithm to evolve the robots. The genetic layer sends the configuration it wants to test to the simulator which returns the fitness value.

## IV. RESULTS

All experiments and results are detailed in this section. Ten experiments of body-brain evolution were initially run to validate the effectiveness of the "TensSoft Evolution" body-brain co-evolution platform and to investigate the effect of different modules stiffness. The objective of the evolution (i.e. fitness function), for all the experiments, was to maximize the distance reached by the robot after 10 seconds.

### A. Experimental conditions

We ran two sets of experiments to compare robotic modules with two different stiffnesses. The two sets were composed of 5 experiments each for a total of 10 experiments. These were all initialized with different random seeds. All other parameters were fixed. Here some of them:
- Range of number of modules: 2-9
- Frequency range of module actuators: 0-1 Hz
- Amplitude range of module actuators: 0-1
- Number of Generations: 200
- Number of Individuals: 50

The previous actuation parameters were chosen to match the servo cabilities present in the real hardware. The only changing parameter between the 2 sets was the modules' stiffness; the HIGH Stiffness = 4 x LOW Stiffness.

Both stiffnesses are in the elastic domain (10-100 MPa) similar to soft materials ones.

### B. Fitness results and evolved locomotion strategies

The table 1 is a summary of all different evolution conditions and results.

The evolution with only stiffer modules favored long robots of 6-8 modules and evolved locomotion strategies similar to that of a caterpillar (CAT) (e.g. activation in sequence observed like in peristaltic locomotion). See attached movie at reference [27].

The evolution with lower stiffness modules consistently selects robots made out of only 2-3 modules and evolves a hopping behavior (Hop) or a rolling on its side behavior (Rol). In the Hop/Rol behavior, one module is compressed to store energy as a spring and one-two additional modules are used to orient the re-expansion of it to propel the robot forward or sideward. See attached movie at reference [28].

TABLE I. EVOLUTION CONDITION AND RESULTS

| Evolution | 1 | 2 | 3 | 4 | 5 | 6 | 7 | 8 | 9 | 10 |
|---|---|---|---|---|---|---|---|---|---|---|
| Stiffness | High | High | High | High | High | Low | Low | Low | Low | Low |
| Best Individual's number of modules | 8 | 8 | 7 | 8 | 6 | 2 | 2 | 3 | 2 | 2 |
| Best Individual's Fitness | 307 | 320 | 450 | 339 | 405 | 136 | 212 | 175 | 172 | 155 |
| Best Individual's Strategy | CAT | CAT | CAT | CAT | CAT | Hop | Hop | Hop/Rol | Hop | Hop/Rol |

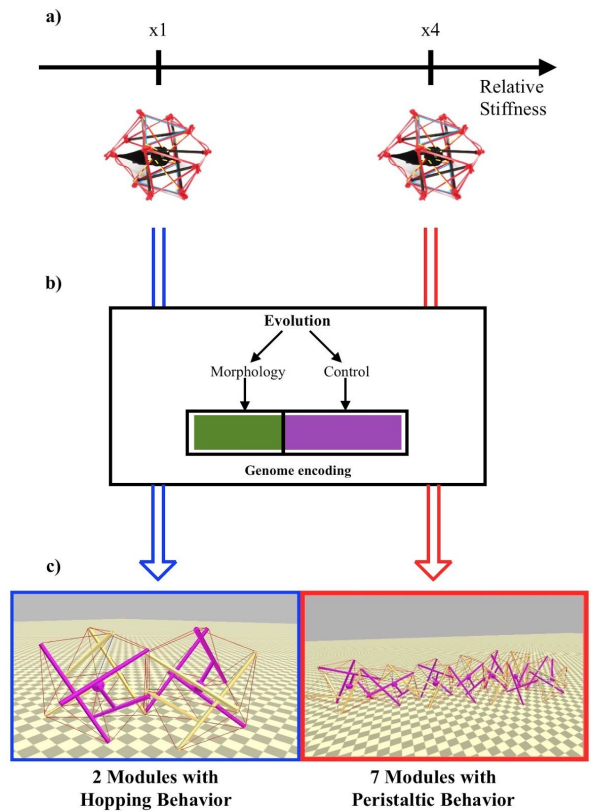

Fig. 3. (a) Stiffness scale of the module. (b) Co-evolution of morphology and control with a sequential genome. (c) Different modules stiffnesses lead to different morphology, control and locomotion strategy.

The different evolutions gave consistently similar results for the same stiffness in both design and control strategy.

Our hypothesis to explain the drastic difference in results between the 2 experiments is that peristaltic locomotion requires a specific stiffness and length of the robot for a certain scale of the tensegrity modular robot, and friction of the ground [29]. This could mean that the low stiffness experiment finds it less optimal to evolve a caterpillar structure. It therefore tries other approaches and finds the hopping as being a more efficient locomotion strategy. This demonstrates how the stiffness of a tensegrity robot is an important design parameter.

V. CONCLUSIONS AND DISCUSSION

In this work, we presented our modular strategy to develop soft tensegrity robots and we showed that co-evolution of body and brain is an effective design strategy. Moreover, we show the importance of designing body stiffness by demonstrating the emergence of completely different locomotion strategies for the same task. An additional result of this work is the development of the "TensSoft" platform for co-evolution of complex tree-structures that also allows the evolution of articulated modular tensegrity robots. More complex morphologies will be the subjects of study in the future work.

Future work will also investigate the possibility of evolving different types of modules with different stiffness in the same robot morphology (e.g. stiffness distribution across robot's body) and how this affects robots' adaptability to different scenarios.

In conclusion, we believe that co-evolution of soft tensegrity robots with programmable stiffness paves the way to an effective design method for highly deformable and robust tensegrity soft robots, suitable for working safely in complex unstructured environments or in human-centric environments.